# Skin Lesion Analysis towards Melanoma Detection Using Deep Learning Network

Yuexiang Li and Linlin Shen*

*Abstract*—Skin lesion is a severe disease in world-wide extent. Early detection of melanoma in dermoscopy images significantly increases the survival rate. However, the accurate recognition of melanoma is extremely challenging due to the following reasons, e.g. low contrast between lesions and skin, visual similarity between melanoma and non-melanoma lesions, etc. Hence, reliable automatic detection of skin tumors is very useful to increase the accuracy and efficiency of pathologists. *International Skin Imaging Collaboration* (ISIC) is a challenge focusing on the automatic analysis of skin lesion. In this paper, we proposed two deep learning methods to address all the three tasks announced in ISIC 2017, i.e. lesion segmentation (task 1), lesion dermoscopic feature extraction (task 2) and lesion classification (task 3). A deep learning framework consisting of two fully-convolutional residual networks (FCRN) is proposed to simultaneously produce the segmentation result and the coarse classification result. A lesion index calculation unit (LICU) is developed to refine the coarse classification results by calculating the distance heat-map. A straight-forward CNN is proposed for the dermoscopic feature extraction task. To our best knowledges, we are not aware of any previous work proposed for this task. The proposed deep learning frameworks were evaluated on the ISIC 2017 testing set. Experimental results show the promising accuracies of our frameworks, i.e. 0.718 for task 1, 0.833 for task 2 and 0.823 for task 3 were achieved.

## I. Introduction

Melanoma is the most deadly form of skin cancer and accounts for about 75% of deaths associated with skin cancer [1]. Accurate recognition of melanoma in early stage can significantly increase the survival rate of patients. However, the manual detection of melanoma produces huge demand of well-trained specialists, and suffers from inter-observer variations. A reliable automatic system for melanoma recognition increasing the accuracy and efficiency of pathologists is worthwhile to develop.

Dermoscopy technique has been developed to improve the diagnostic performance of melanoma. Dermoscopy is a noninvasive skin imaging technique of acquiring a magnified and illuminated image of skin region for increased clarity of the spots [2], which enhances the visual effect of skin lesion by removing surface reflection. Nevertheless, automatic recognition of melanoma from dermoscopy images is still a difficult task as it has several challenges. First, the low contrast between skin lesions and normal skin region makes it difficult to segment accurate lesion areas. Second, the melanoma and non-melanoma lesions may have high degree of visual similarity, resulting in the difficulty for distinguishing melanoma lesion from non-melanoma. Third, the variation of skin conditions, e.g. skin color, natural hairs or veins, among patients produce different appearance of melanoma, in terms of color and texture, etc.

Skin lesion segmentation is the essential step for most classification approaches. Recent review of automated skin lesion segmentation algorithms can be found in [3, 4]. Accurate segmentation can benefit the accuracy of subsequent lesion classification. Extensive studies have been made to produce decent lesion segmentation results. For example, Gomez et al. proposed an unsupervised algorithm, named Independent Histogram Pursuit (IHP), for the segmentation of skin lesion [5]. The algorithm was tested on five different dermatological datasets and achieved a competitive accuracy close to 97%. Garnavi et al. proposed an automated segmentation approach for skin lesion using optimal color channels and hybrid thresholding technique [6]. In more recent research, Pennisi et al. employed Delaunay Triangulation to extract binary mask of skin lesion region, which does not require any training stage [7]. Yu used deep learning approach, i.e. fully-convolutional residual network (FCRN), for skin lesion segmentation in dermoscopy images [8] and achieved a competitive result on ISIC 2016 dataset.

Based on the segmentation results, hand-crafted features can be extracted for melanoma recognition. Celebi et al. extracted several features including color and texture from segmented lesion region for skin lesion classification [9]. Schaefer used automatic border detection approach [10] to segment lesion area and then ensemble the extracted features, i.e. shape, texture and color, for melanoma recognition [11]. On the other hand, some investigations [12-14] attempted to directly employ hand-crafted feature for melanoma recognition without segmentation step. Different from approaches using hand-crafted features, deep learning network uses hierarchical structure to automatically extract features. Due to the breakthroughs made by deep learning in increasing number of medical image processing tasks, some research started to apply deep learning approach for melanoma recognition. Codella et al. proposed a hybrid approach integrating convolutional neural

The work was supported by Natural Science Foundation of China under grands no. 61672357, the Science Foundation of Shenzhen under Grant No. JCYJ20160422144110140. Corresponding author: Prof. Linlin Shen

Yuexiang Li and Linlin Shen are with Computer Vision Institute, College of Computer Science and Software Engineering, Shenzhen University, Shenzhen, China. (Email: yuexiang.li@szu.edu.cn, llshen@szu.edu.cn)

network (CNN), sparse coding and support vector machine (SVM) to detect melanoma [15]. In the recent research, Codella and his colleagues established a system combining recent developments in deep learning and machine learning approaches for skin lesion segmentation and classification [16]. Kawahara et al. employed a fully convolutional network to extract multi-scale features for melanoma recognition [17]. Yu et al. applied very deep residual network to distinguish melanoma from non-melanoma lesions [8].

Although lots of work was proposed, there is still margin of performance improvement for both skin lesion segmentation and classification. *International Skin Imaging Collaboration* (ISIC) continuously organized melanoma detection challenges from 2016, which highly promotes the accuracy of automatic melanoma detection methods. In ISIC 2017, three processing tasks of skin lesion images including lesion segmentation, dermoscopic feature extraction and lesion classification, were announced. Different from the lesion segmentation and classification, which have been extensively studied, dermoscopic feature extraction is a new task in the area. Consequently, few of study are proposed to address the problem.

The main contribution of this paper can be summarized as follows:

1) Existing deep learning approaches commonly use two networks to separately perform lesion segmentation and classification. In this paper, we proposed a framework consisting of multi-scale fully-convolutional residual networks and a lesion index calculation unit (LICU) to simultaneously address lesion segmentation (task 1) and lesion classification (task 3). The proposed framework achieved comparable results to the state-of-the-art in both tasks. Henceforth, the proposed framework is named as Lesion Indexing Network (LIN).
2) We proposed a CNN-based framework, named Lesion Feature Network (LFN), to address task 2, i.e. dermoscopic feature extraction. Experimental results demonstrate the competitive performance of our framework. To our best knowledge, we are not aware of any previous work proposed for this task. Hence, this work may become the benchmark for the following related research in the area.
3) We made detailed analysis of the proposed deep learning frameworks in several aspects, e.g. the performances of networks with different depths; the influences caused by adding different components (e.g. batch normalization, weighted softmax, etc.). This work provides useful guidelines for the design of deep learning network in related medical research.

## II. Methods

In this section, we introduce the deep learning methods developed for different tasks.

### A. Lesion segmentation and classification (task 1 & 3)

*1) Pre-processing*

The original ISIC skin lesion dataset contains 2000 images of different resolutions. The resolutions of some lesion images are above 1000 x 700, which require high cost of computation. It is necessary to rescale the lesion images for deep learning network. As directly resizing image may distort the shape of skin lesion, we first cropped the center area of lesion image and then proportionally resize the area to a lower resolution. As illustrated in Fig. 1, this approach not only enlarges the lesion area for feature detection, but also maintains the shape of skin lesion.

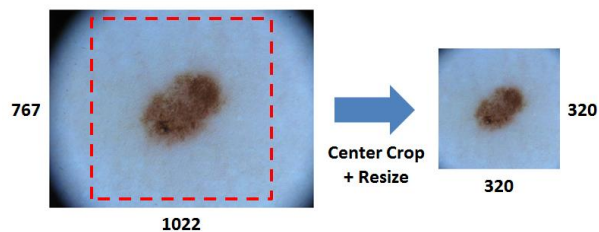

Fig. 1  Pre-processing for skin lesion image. First crop the center area and then proportionally resize to a lower resolution.

*2) Data Augmentation*

As the image volumes of different categories vary widely, we accordingly rotated the images belonging to different classes to establish a class-balanced dataset. The dataset augmented with this step is denoted as *DR*. The image volumes of original training set and DR are listed in Table I. The numbers in the brackets after category names are the angles for each rotation.

TABLE I
DETAILED INFORMATION OF DATA AUGMENTATION (TASK 1 & 3)

|  | **Melanoma (18)** | **Seborrheic Keratosis (18)** | **Nevus (45)** |
|---|---|---|---|
| **Original** | 374 | 254 | 1372 |
| **DR** | 7480 | 5080 | 10976 |

The images in DR are randomly flipped along x or y-axis to establish another pair dataset, called *DM*. The two datasets are used to train FCRNs, respectively.

*3) Lesion Indexing Network (LIN)*
*a) Network Architecture*

The fully convolutional residual network, i.e. FCRN-88, proposed in our previous work [18], which outperforms the FCRN-50 and FCRN-101 [19], was extended to simultaneously address the tasks of lesion segmentation and classification in this paper. Based on FCRN-88, we construct a Lesion Indexing Network (LIN) for skin lesion image analysis. The flowchart of LIN is presented in Fig. 2. Two FCRNs trained with datasets using different data augmentation methods are involved. The lesion index calculation unit (LICU) is designed to refine the probabilities for *Melanoma*, *Seborrheic keratosis* and *Nevus*.

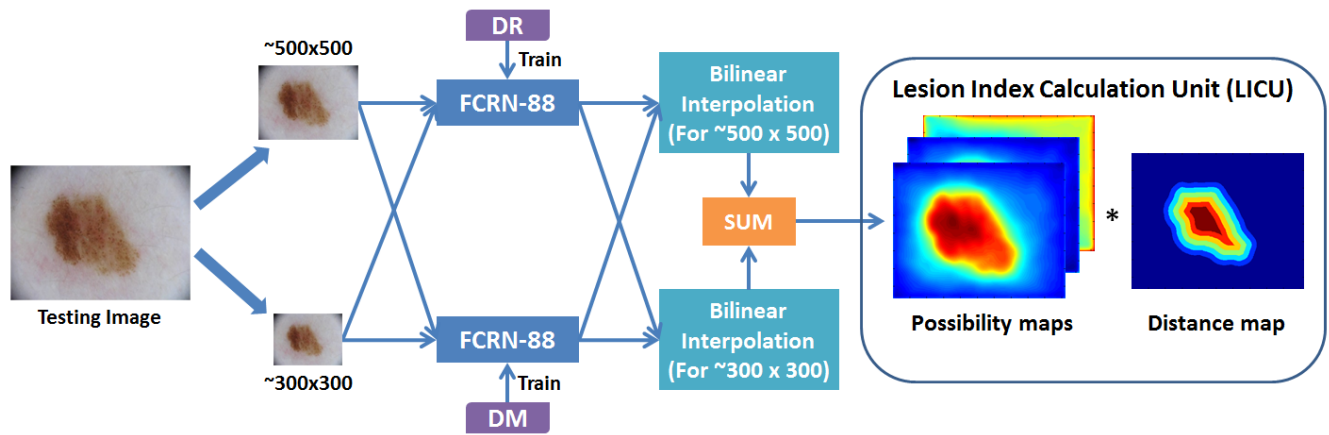

Fig. 2 Flowchart of Lesion Indexing Network (LIN). The framework contains two FCRN and a calculation unit for lesion index.

In the testing stage, as fully-convolutional network accepts inputs with different sizes, we proportionally resize the skin lesion images to two scales, i.e. ~300x300 and ~500x500, and send them to the FCRNs, respectively. The results of different scales are interpolated to the original resolution of testing image and sum up to yield the coarse possibility maps. The LICU employs a distance map representing the importance of each pixel to refine the coarse possibilities of skin lesions.

The reason for using separate FCRN-88 trained on different datasets, i.e. DR and DM, is that we found 'mirror' operation seems to fool the FCRN-88 during training. In our experiments, single FCRN-88 trained on the combination of DR and DM was easy to be overfitting. The segmentation and classification accuracies on validation set verified our findings, i.e. the separate network provides better segmentation and classification performance than that of single FCRN-88 trained using DR+DM.

*b) Lesion Index Calculation Unit (LICU)*

As the accurate possibility maps of different lesion categories of skin lesion image provide useful information for pathologists, we proposed a component, named Lesion Index Calculation Unit (LICU), to refine the coarse skin lesion possibilities maps from FCRNs.

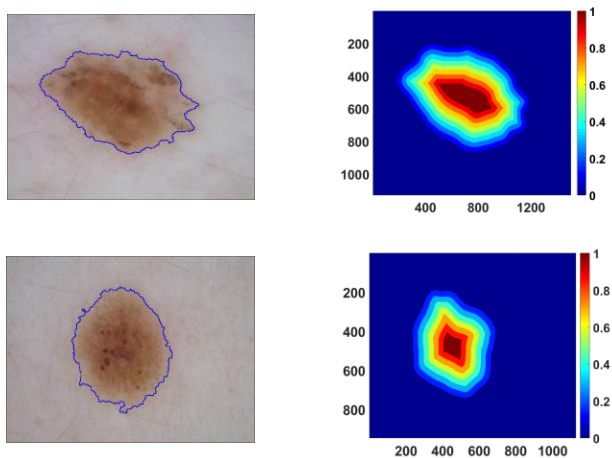

Fig. 3 Examples of skin lesion images with outlines (blue) and distance maps. The first column is the original lesion images and the second is the corresponding distance maps.

First, the coarse possibilities maps after summation need to be normalized to [0, 1]. Let $v_i(x, y)$ be the value of $(x, y)$ in $i^{th}$ coarse map, the normalized possibility for skin lesions $(p_i)$ can be deduced by:

$$p_i(x,y) = \frac{v_i(x,y) - \min_{i \in 1,2,3}(v_i(x,y))}{\sum_{i=1}^{3}(v_i(x,y) - \min_{i \in 1,2,3}(v_i(x,y)))} \quad i \in 1,2,3 \quad (1)$$

Each pixel in the lesion area has different importance for lesion classification. It can be observed from Fig. 3 that the area near lesion border of some skin lesion images has more similar appearance, i.e. color/texture, to skin than that of the center area. The blue lines in Fig. 3 are the borders of lesions produced by LIN. The lesion area with similar features to skin may provide less information for lesion recognition. Hence, the distances from pixels to the nearest border are used to represent the importance of pixels for lesion classification. Examples of distance maps are shown in the second column of Fig. 3. The colors in distance map represent the weights for corresponding pixels. The distance map is multiplied to each of the normalized coarse possibilities maps to generate refined maps. Finally, we average the possibilities in the lesion area of refined maps to obtain the indexes for different categories of skin lesion.

*4) Implementation*

The proposed LIN is established using *MatConvNet* toolbox [20]. While 80% of the training dataset is used for training, the remaining is used for validation. The FCRNs were individually trained with a mini-batch size of 128 on one GPU (GeForce GTX TITAN X, 12GB RAM). The details of training setting are the same to [18]. The network converges after 6 epochs of training.

B. *Dermoscopic feature extraction (task 2)*

The dermoscopic feature extraction is a new task announced in ISIC 2017, which aims to extract clinical features from dermoscopic images. Few of previous work were proposed for this task. In this section, we introduce a CNN-based approach, i.e. Lesion Feature Network (LFN), developed to address the challenge.

*1) Superpixels extraction*

The ISIC dermoscopic images were subdivided into superpixels using algorithm introduced in [21]. An example is shown in Fig. 4. The original skin lesion image (Fig. 4 (a)) was divided to 996 superpixel areas (Fig. 4 (b)), which are separated by black lines.

The superpixel masks contain four kinds of dermoscopic features, i.e. Pigment Network (**PN**), Negative Network (**NN**), Streaks (**S**) and Milia-like Cysts (**MC**), and background (**B**). We extract the content of each superpixel according to [21] and resize them to a uniform size, i.e. 56x56, for the proposed Lesion Feature Network.

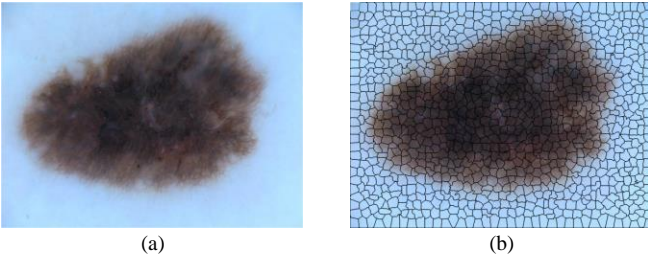

(a) (b)
Fig. 4 Example of superpixels. The original image (a) was subdivided to 996 superpixel areas (b) separated by black lines.

*2) Data augmentation*

The extracted patch dataset is extremely imbalanced. Most of patches only contain the background information. Hence, data augmentation processing is needed to balance the image volumes of different categories. Two processing techniques, i.e. Random sample & Patch rotation, were adopted. The image volumes of original and augmented patch datasets are listed in Table II.

TABLE II
DETAILED INFORMATION OF DATA AUGMENTATION (TASK 2)

|  | Original | Random sample + Rotation |
|---|---|---|
| **Background (B)** | >> 90,000 | 87,089 |
| **Pigment Network (PN)** | > 80,000 | 77,325 |
| **Negative Network (NN)** | ~3,000 | 12,908 |
| **Milia-like Cysts (MC)** | ~5,000 | 18,424 |
| **Streaks (S)** | ~2,000 | 8,324 |

*a) Random sample*

As listed in Table II, the volume of original background patches is much larger than that of other categories. However, most of background patches contain similar contents. Hence, background patches have lots of redundant information. To remove the redundancy and decrease patch volume, the background patches for LFN training are randomly selected from original patch dataset, which finally formed a set of 87,089 background patches.

Due to the extremely large volume of Pigment Network (PN) in original patch dataset, random sample operation was also applied to PN, which results in a set of 77,325 PN patches.

*b) Patch rotation*

The volumes of NN, MC and S patches are relative small in original dataset. Image rotation is employed to augment the volumes. Three angles, i.e. 90, 180 and 270, were adopted for patch rotation, which increases the patch volumes to 12,908, 18,424 and 8,324 for NN, MC and S, respectively.

*3) Lesion Feature Network (LFN)*

The augmented training set was used to train our Lesion Feature Network (LFN), whose architecture is presented in Fig. 5.

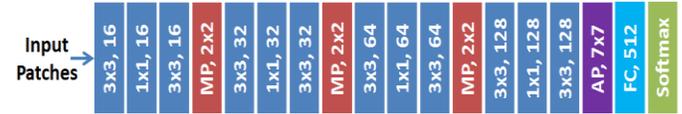

Fig. 5 Flowchart of Lesion Feature Network (LFN).

While the blue rectangles represent the convolutional layers, the numbers represent kernel size and number of kernels. LFN involves 12 convolutional layers for feature extraction, which can be separated to 4 stages, i.e. 3 convolutional layers per stage. As the 1x1 convolution can integrate the features extracted by 3x3 convolution for better feature representation, a network in network like structure [22] is adopted for each stage. FC is the fully-connected layer. Both max pooling (MP) and average pooling (AP) are used and the network was trained with softmax loss, defined in (2).

$$L = \frac{1}{N}\sum_i L_i = \frac{1}{N}\sum_i -\log\left(\frac{e^{f_{y_i}}}{\sum_j e^{f_j}}\right) \qquad (2)$$

where $f_j$ denotes the *j*-th element ($j \in [1, K]$, *K* is the number of classes) of vector of class scores $f$, $y_i$ is the label of *i*-th input feature and *N* is the number of training data.

Although the data augmentation operation was performed, the obtained training dataset is still imbalanced. To address the problem, weights are assigned for different classes while calculating the softmax loss, to pay more attentions to the classes with fewer samples. According to the image volumes in augmented training set, the weights are set to 1, 1, 5, 3 and 8 for **B**, **PN**, **NN**, **MC** and **S**, respectively.

*4) Implementation*

The proposed LFN is developed using *Keras* toolbox. The patch dataset is separated to training set and validation set according to the percentages of 80:20. The network is optimized by Stochastic Gradient Descent (SGD) [23] with an initial learning rate of 0.01 and a momentum of 0.9. The learning rate decreases with gamma = 0.1. The network was trained on a single GPU (GeForce GTX TITAN X, 12GB RAM) and was observed to converge after 10 epochs of training.

III. PERFORMANCE ANALYSIS

*A. Datasets*

*International Skin Imaging Collaboration* (ISIC) continuously organized melanoma detection challenges from 2016. In this year, the organizer provides larger image data set for lesion segmentation & classification and adds the annotations for dermoscopic feature extraction. ISIC 2017 provides 2000 skin lesion images as training set with masks for segmentation, superpixel masks for dermoscopic feature extraction and annotations for classification. The lesion images are classified to three categories, Melanoma, Seborrheic keratosis and Nevus. Melanoma is the malignant skin tumor, which leads to high death rate. The other two kinds of lesion, i.e. Seborrheic keratosis and Nevus, are the benign skin tumor derived from different cells. Fig. 6 presents the lesion images from ISIC 2017 and their masks for different tasks. The first row in Fig. 6 is the original skin lesion images. The second row is the masks for lesion segmentation, while the third row is the superpixel masks for dermoscopic feature extraction.

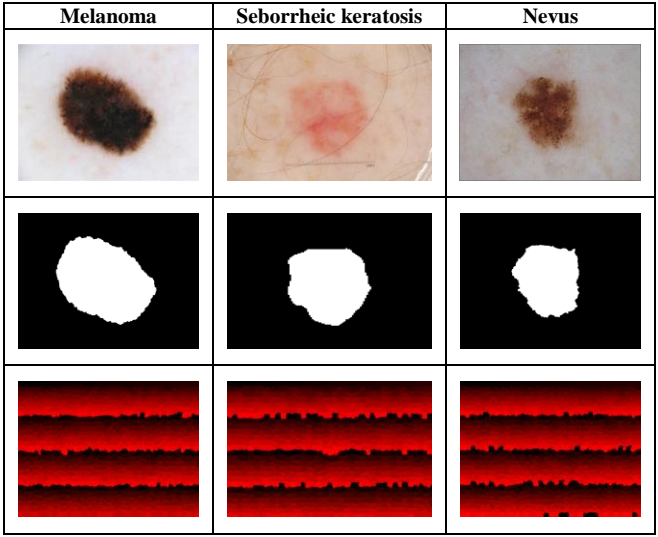

Fig. 6 Examples of lesion images from ISIC 2017 and their masks. The first row is the original images of different lesions. The second row is the segmentation masks. The third row is the superpixel mask for dermoscopic feature extraction.

ISIC 2017 also provides a validation set of 150 images for participants to evaluate their framework. The final competition is made on the ISIC 2017 testing set, which consists of 600 skin lesion images. As the ground truths of validation and testing sets are unpublished during the competition, participants need to submit their results to the online system for assessment. Recently, the ISIC 2017 organizers released the ground truths for validation and testing set[1].

In this section, we analyze the performances of proposed LIN and LFN on ISIC 2017 validation set. The comparison with state-of-the-art on ISIC 2017 testing set will be presented in the next section.

### B. Evaluation Metrics

*1) Lesion segmentation*

The challenge employs several metrics for performance evaluation, which includes accuracy (AC), Jaccard Index (JA), Dice coefficient (DI), sensitivity (SE) and specificity (SP). Let $N_{tp}$, $N_{tn}$, $N_{fp}$ and $N_{fn}$ represent the number of true positive, true negative, false positive and false negative, respectively, the criteria can be defined as:

[1] https://challenge.kitware.com/#challenge/

$$AC = \frac{N_{tp}+N_{tn}}{N_{tp}+N_{fp}+N_{tn}+N_{fn}}, \quad (3)$$

$$JA = \frac{N_{tp}}{N_{tp}+N_{fn}+N_{fp}}, \quad DI = \frac{2*N_{tp}}{2*N_{tp}+N_{fn}+N_{fp}}, \quad (4)$$

$$SE = \frac{N_{tp}}{N_{tp}+N_{fn}}, \quad SP = \frac{N_{tn}}{N_{tn}+N_{fp}} \quad (5)$$

The segmentation results from participants are ranked by the JA metric. The other metrics are measured as reference in the challenge.

*2) Dermoscopic feature extraction & Lesion classification*

The same evaluation metrics, i.e. AC, SE and SP, are employed to assess the performance of dermoscopic feature extraction and lesion classification. Average precision (AP) defined in [24] is also involved. The primary metric ranking the results for these two tasks is the area under the ROC curve, i.e. AUC, which is generated by evaluating the true positive rate (TPR), i.e. SE, against the false positive rate (FPR), defined in (6), at various threshold settings.

$$FPR = \frac{N_{fp}}{N_{tn}+N_{fp}} = 1 - SP \quad (6)$$

### C. Lesion Indexing Network (LIN)

*1) The performance on lesion segmentation*

To visually analyze the segmentation performance of proposed LIN, some examples of its segmentation results are presented in Fig. 8. The blue and red lines represent the segmentation outlines of LIN and the ground truths, respectively. The examples illustrate some primary challenges in the area of skin lesion image processing. The contrast between lesion and skin region is low in Fig. 8 (b), (c) and (f). Human hair near the lesion region of Fig. 8 (d) may influence the segmentation. Nevertheless, it can be observed from Fig. 8 that the proposed Lesion Indexing Network yields satisfied segmentation results for all of the challenging cases.

*a) Training with DR and DM*

In the experiments, 'rotation and 'mirror' operations were adopted to enlarge the training dataset for Lesion Indexing Network. However, the FCRN-88 seems to be fooled by the 'mirror' operation. Fig. 7 shows the loss curves of FCRN-88 trained with DR, DM and DR+DM, respectively.

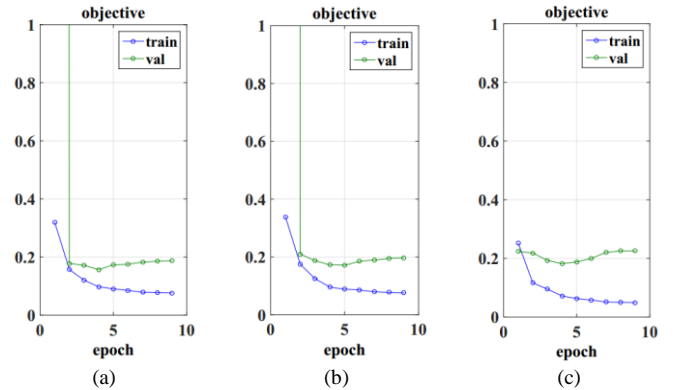

Fig. 7 Loss curves of LIN trained with DR (a), DM (b) and DR+DM (c).

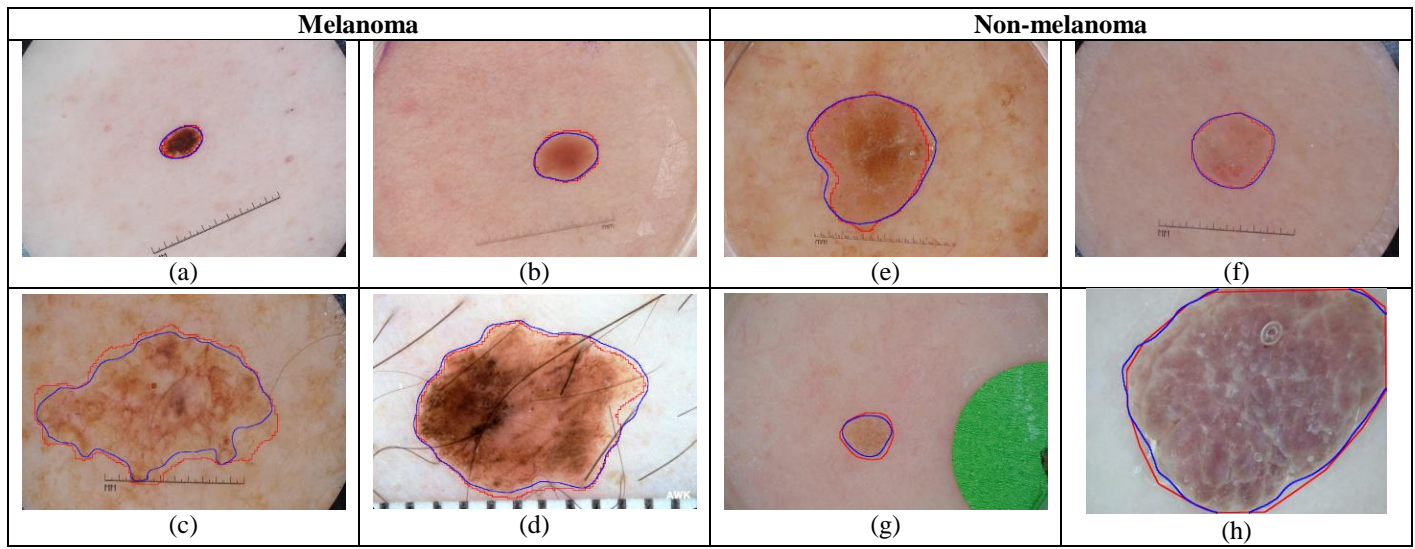

Fig. 8 Examples of skin lesion segmentation results produced by LIN for ISIC 2017 validation set. (a)-(b) are the results of Melanoma while (e)-(f) are the results for Seborrheic keratosis and Nevus. The blue and red lines represent the segmentation results and ground truths.

The validation loss of FCRN-88 trained on DR/DM is stable around 0.2. In contrast, the loss of FCRN-88 trained by DR+DM decreases to about 0.18 and then gradually increases to over 0.2. The FCRN-88 trained with DR+DM has the lowest training loss but the highest validation loss among the frameworks. It is because the samples of DR and DM are paired. The similar appearances of paired samples make the very deep FCRN-88 easily overfitting to the dataset.

Table III listed the JA of single FCRN-88 trained on DR/DR+DM and our LIN evaluated on ISIC 2017 validation set. For comparison convenience, the frameworks only take single scale of lesion images, i.e. ~300x300, as input. As shown in Table III, due to the overfitting problem, the JA of FCRN-88 trained with DR+DM is the lowest, i.e. 0.607. The proposed LIN achieves the best performance, i.e. 0.710.

TABLE III
JA OF FRAMEWORKS ON ISIC 2017 VALIDATION SET

| Model | JA |
| --- | --- |
| FCRN-88 (DR) | 0.697 |
| FCRN-88 (DR+DM) | 0.607 |
| LIN (ours) | **0.710** |

*b)  Experiments on the multi-scale input images*

Taking the computation efficiency into account, the original skin lesion images were cropped and resized to 320x320 for network training. However, lesion images of larger scale (~500x500) provide a clearer view of lesion area, e.g. the texture, for feature extraction. To demonstrate the importance of processing skin lesion image at multi-scales, a set of experiments were conducted. Three scales of testing images were selected, i.e. ~300x300, ~500x500 and ~700x700, for comparison.

TABLE IV
JA OF FRAMEWORKS WITH DIFFERENT SCALES OF INPUTS

| Model | JA |
| --- | --- |
| LIN (~300) | 0.710 |
| LIN (~500) | 0.698 |
| LIN (~700) | 0.662 |
| LIN (~300 + ~500) | 0.751 |
| LIN (~300 + ~500 + ~700) | **0.753** |

For single scale, input image of ~300 achieves the best performance on ISIC validation set, i.e. a JA of 0.710. Degradation of segmentation performance is observed when only using the larger-scale images, i.e. degradations of 0.012 and 0.048 for ~500 and ~700, respectively. However, the larger-scale input images can assist LIN to perform more decent segmentation. The LIN using all of three scales achieves the best JA, i.e. 0.753, which is 0.002 higher than the second-rank, i.e. LIN using ~300 and ~500. In consideration of computational efficiency, the LIN using ~300 and ~500 is preferable for experiments and applications.

*2)  The performance on lesion classification*
*a)  Performance of LICU*

Each pixel in the lesion images has different importance for the final classification result. Although the FCRN-88 can simultaneously perform segmentation and classification tasks, it assigns equal importance for all pixels. Lesion Index Calculation Unit (LICU) measures the pixel importance by distance map and accordingly refines the possibility maps from FCRN-88s. Experiments were conducted on ISIC 2017 validation set to assess the performance of LICU. Table V lists the results. Compared to the plain LIN, i.e. 0.891, the LICU component produces an improvement of 0.017 for LIN, i.e. 0.908.

TABLE V
AUC OF FRAMEWORKS WITH/WITHOUT LICU

| Model | AUC |
| --- | --- |
| LIN without LICU | 0.891 |
| LIN with LICU | **0.908** |

*D.  Lesion Feature Network (LFN)*

*1)  Analysis of network architecture*

To analyze the influence caused by layer width, we transform the original LFN to two variations for comparison, i.e. Narrow LFN and Wide LFN, whose detailed information is listed in Table VI.

TABLE VI
DETAILED INFORMATION OF DIFFERENT LFNS

|  | LFN | Narrow LFN | Wide LFN |
|---|---|---|---|
| **Stage 1** | 16, (3,3)<br>16, (1,1)<br>16, (3,3) | 16, (3,3)<br>16, (1,1)<br>16, (3,3) | 32, (3,3)<br>32, (1,1)<br>32, (3,3) |
| **Stage 2** | 32, (3,3)<br>32, (1,1)<br>32, (3,3) | 16, (3,3)<br>16, (1,1)<br>16, (3,3) | 64, (3,3)<br>64, (1,1)<br>64, (3,3) |
| **Stage 3** | 64, (3,3)<br>64, (1,1)<br>64, (3,3) | 16, (3,3)<br>16, (1,1)<br>16, (3,3) | 64, (3,3)<br>64, (1,1)<br>64, (3,3) |
| **Stage 4** | 128, (3,3)<br>128, (1,1)<br>128, (3,3) | 32, (3,3)<br>32, (1,1)<br>32, (3,3) | 128, (3,3)<br>128, (1,1)<br>128, (3,3) |

The performances of three LFNs were evaluated on ISIC 2017 validation set in Table VII. By comparing the AUC of LFN and Narrow LFN, we notice that the narrow layer decreases the capacity of feature representation of framework. The AUC of Narrow LFN is 0.822, which is 0.026 lower than that of LFN, i.e. 0.848. In another aspect, too wide layer leads to the overfitting problem, which also decreases the performance of LFN. The AUC of wide LFN (0.803) is 0.045 lower than that of original LFN. Hence, the proposed LFN better balance the relationship between feature representation capacity of framework and network overfitting problem.

*2) Performance of weighted softmax loss (WSL)*

Although data augmentation approach was used to balance the sample volumes of different categories, the generated training set is still imbalanced. Weighted softmax loss (WSL) is another important tool to alleviate the influence caused by imbalanced training set during network training. As shown in Table VII, without using WSL, the AUC of LFN sharply decreases to 0.778, which demonstrates the importance of weighted softmax loss.

*3) Usage of Batch Normalization (BN)*

Batch normalization (BN) [25] component can reduce internal covariate shift and accelerate the training process, which has been widely adopted in many deep learning frameworks, e.g. ResNet [19] and Inception [26]. In the proposed LFN, BN is adopted between convolutional layer and rectified linear units layer. The result presented in Table VII indicates that an improvement of 0.006 is generated by BN component for AUC.

TABLE VII
AUC OF LFNS ON VALIDATION SET

| Model | AUC |
|---|---|
| Narrow LFN | 0.822 |
| Wide LFN | 0.803 |
| LFN | **0.848** |
| LFN (without WSL) | 0.778 |
| LFN (without BN) | 0.842 |

IV. COMPARISON WITH STATE-OF-THE-ART

We participated all of the three tasks announced in ISIC 2017 challenges. In the testing stage, there are totally 21 submissions for lesion segmentation (task 1), 3 submissions for dermoscopic feature extraction (task 2) and 23 submissions for lesion classification (task 3). The results were evaluated using the above mentioned metrics. The final rank of lesion segmentation task is based on JA. AUC is the metric for the final rank of task 2 and 3. We ranked the ninth place, i.e. a JA of 0.718, in task 1, second place, i.e. an AUC of 0.833, in task 2 and tie for the twelfth place, i.e. an AUC of 0.823, in task 3. The top-15 results of different tasks were listed in Table VIII, Table IX and Table X, respectively.

The JA and AUC of our LIN for lesion segmentation and classification are 0.718 and 0.823, which are comparable to the competition winners. The proposed LFN achieves the best average precision (AP) and sensitivity (SE), i.e. 0.409 and 0.665, for the dermoscopic feature extraction task, which are 0.168 and 0.123 higher than that of the participant ranking the second place.

TABLE VIII
RESULTS OF SKIN LESION SEGMENTATION ON ISIC 2017

| Method | JA | AC | DC | SE | SP |
|---|---|---|---|---|---|
| Mt.sinai | **0.765** | **0.934** | **0.849** | 0.825 | 0.975 |
| NLP LOGIX | 0.762 | 0.932 | 0.847 | 0.820 | 0.978 |
| USYD-BMIT | 0.760 | **0.934** | 0.844 | 0.802 | **0.985** |
| USYD-BMIT | 0.758 | **0.934** | 0.842 | 0.801 | 0.984 |
| RECOD Titans | 0.754 | 0.931 | 0.839 | 0.817 | 0.970 |
| Jeremy Kawahara | 0.752 | 0.930 | 0.837 | 0.813 | 0.976 |
| NedMos | 0.749 | 0.930 | 0.839 | 0.810 | 0.981 |
| INESC TEC Porto | 0.735 | 0.922 | 0.824 | 0.813 | 0.968 |
| **LIN (ours)** | 0.718 | 0.922 | 0.810 | 0.789 | 0.975 |
| GGAMA | 0.715 | 0.915 | 0.797 | 0.774 | 0.970 |
| naiven | 0.697 | 0.910 | 0.795 | 0.790 | 0.982 |
| UESTC-JQI | 0.684 | 0.917 | 0.774 | 0.784 | 0.950 |
| DAnMI | 0.679 | 0.900 | 0.774 | 0.779 | 0.962 |
| eVida | 0.665 | 0.884 | 0.760 | **0.869** | 0.923 |
| Yale | 0.665 | 0.910 | 0.775 | 0.812 | 0.951 |

TABLE IX
RESULTS OF DERMOSCOPIC FEATURE EXTRACTION ON ISIC 2017

| Method | AUC | AC | AP | SE | SP |
|---|---|---|---|---|---|
| Jeremy Kawahara | **0.895** | **0.980** | 0.241 | 0.542 | **0.981** |
| **LFN (ours)** | 0.833 | 0.914 | **0.409** | **0.665** | 0.915 |

TABLE X
RESULTS OF SKIN LESION CLASSIFICATION ON ISIC 2017

| Method | AUC | AC | AP | SE | SP |
|---|---|---|---|---|---|
| CSUJT | **0.911** | 0.816 | 0.748 | **0.856** | 0.812 |
| MPG-UCIIIM | 0.910 | 0.849 | 0.747 | 0.140 | **0.998** |
| RECOD Titans | 0.908 | 0.883 | **0.752** | 0.451 | 0.970 |
| USYD-BMIT | 0.896 | **0.888** | 0.732 | 0.508 | 0.970 |
| IHPC-NSC | 0.886 | 0.873 | 0.665 | 0.568 | 0.940 |
| UoG-MLRG | 0.886 | 0.879 | 0.703 | 0.453 | 0.971 |
| icuff | 0.851 | 0.819 | 0.578 | 0.524 | 0.893 |
| icuff | 0.850 | 0.817 | 0.579 | 0.524 | 0.890 |
| USYD-BMIT | 0.836 | 0.850 | 0.569 | 0.210 | 0.989 |
| CVI | 0.829 | 0.863 | 0.593 | 0.460 | 0.950 |
| UoD | 0.825 | 0.849 | 0.557 | 0.591 | 0.907 |
| INESC TEC Porto | 0.823 | 0.657 | 0.566 | 0.814 | 0.615 |
| UFMG | 0.823 | 0.830 | 0.567 | 0.488 | 0.900 |
| **LIN (ours)** | 0.823 | 0.852 | 0.476 | 0.504 | 0.930 |
| IPA | 0.811 | 0.811 | 0.542 | 0.362 | 0.901 |

V. CONCLUSION

In this paper, we proposed two deep learning frameworks, i.e. Lesion Indexing Network (LIN) and Lesion Feature Network

(LFN), to address three primary challenges of skin lesion image processing, i.e. lesion segmentation, dermoscopic feature extraction and lesion classification.

Lesion Indexing Network is proposed to simultaneously address the lesion segmentation and classification. Two very deep fully-convolutional residual networks, i.e. FCRN-88, trained with different training set are adopted to produce the segmentation result and coarse classification result. A lesion indexing calculation unit (LICU) is proposed to measure the importance of pixel for the decision of lesion classification. The coarse classification result is refined according to the distance map generated by LICU.

Lesion Feature Network is proposed to address the task of dermoscopic feature extraction, which is a CNN-based framework trained by the patches extracted from the superpixel masks. To our best knowledge, we are not aware of any previous work available for this task. Hence, this work may become a benchmark for following related research.

Our deep learning frameworks have been evaluated on the ISIC 2017 testing set. The JA and AUC of LIN for lesion segmentation and classification are 0.718 and 0.823, which are comparable to the competition winners. The proposed LFN achieves the best average precision and sensitivity, i.e. 0.409 and 0.665 for dermoscopic feature extraction, which demonstrates its excellent capacity addressing the challenge.